%% file: main_arr.tex
\pdfoutput=1

\documentclass[11pt]{article}

\usepackage{ACL2023}

\usepackage{times}
\usepackage{latexsym}
\usepackage{graphicx}
\usepackage{array}
\usepackage{booktabs}
\usepackage{graphicx, amsmath}
\usepackage{adjustbox}
\usepackage{caption}
\usepackage{multirow}
\usepackage{subcaption}
\usepackage[table,xcdraw]{xcolor}
\usepackage{subcaption}     
\usepackage{microtype}
\usepackage{hyperref}
\usepackage{url}
\usepackage{amssymb}
\usepackage{mathtools}
\usepackage{amsthm}
\usepackage{colortbl}
\usepackage{siunitx}
\usepackage{algorithm}
\usepackage{algpseudocode}
\usepackage{float}
\usepackage{wrapfig}

\usepackage{lineno}
\definecolor{darkblue}{rgb}{0, 0, 0.5}
\hypersetup{colorlinks=true, citecolor=darkblue, linkcolor=darkblue, urlcolor=darkblue}

\usepackage[T1]{fontenc}

\usepackage[utf8]{inputenc}

\usepackage{microtype}

\usepackage{inconsolata}
\usepackage{bm}
\usepackage{amsmath}
\usepackage[bb=px]{mathalfa} 
\usepackage[T1]{fontenc}
\usepackage{xspace}

\newcommand{\B}{\bm{B}} 
\newcommand{\G}{\bm{G}}
\newcommand{\SAGC}{\textbf{\textsc{SAGC}}\xspace}
\newcommand{\EB}{\bm{\mathnormal{EB}}}

\title{Faster Synchronous On-Policy RL via Straggler-Aware Group Sizing}

\author{
    \textbf{
    Azal Ahmad Khan\textsuperscript{1,*}\hspace{2em}
    Ammar Ahmed\textsuperscript{1,*}\hspace{2em}
    Zeshan Fayyaz\textsuperscript{2}
    } \\
    \textbf{
    Sheng Di\textsuperscript{3}\hspace{2em}
    Mingyi Hong\textsuperscript{1}\hspace{2em}
    Ali Anwar\textsuperscript{1}
    } \\
    \textsuperscript{1}University of Minnesota\hspace{2em} 
    \textsuperscript{2}University of Waterloo\hspace{2em}
    \textsuperscript{3}Argonne National Laboratory\\
    \textsuperscript{*}Equal contributions\\
    \texttt{\{khan1069, ahme0599, mhong, aanwar\}@umn.edu} \\ \texttt{z3fayyaz@uwaterloo.ca} \hspace{2em} \texttt{sdi1@anl.gov}\\
}

\begin{document}
\maketitle
\begin{abstract}
Synchronous reinforcement learning methods such as Group Relative Policy Optimization (GRPO) provide stable and reproducible on-policy training, but they are highly vulnerable to stragglers, a single unusually long rollout can delay reward computation and parameter updates for the entire group.
This problem becomes more severe as group size increases, creating a tension between the benefits of larger groups and the wall-clock cost of synchronization stalls.
We propose \underline{S}traggler-\underline{A}ware \underline{G}roup \underline{C}ontrol (\SAGC), a dynamic group-size controller that adapts the training group online based on observed rollout behavior.
\SAGC formulates group-size selection as an online constrained optimization problem, seeking to retain the benefits of larger groups while controlling the long-term rate of straggler events. Across synchronous GRPO and DAPO training, and on top of both vanilla and strong engineered baselines, \SAGC consistently reduces straggler incidence and improves wall-clock efficiency while achieving competitive or better training reward.
We further show that these gains transfer to final model quality: \SAGC is competitive with or better than the strongest static group-size baseline on downstream reasoning benchmarks, and often produces shorter outputs without any explicit length penalty. These results position dynamic group control as a practical way to make synchronous on-policy RL more efficient and robust.
\end{abstract}

\input{sections/introduction}

\input{sections/related_work}
\input{sections/motivation}
\input{sections/method}
\input{sections/experiments}
\input{sections/conclusion}
\input{sections/limitations}

\bibliography{bibliography}
\bibliographystyle{acl_natbib}  

\newpage
\input{sections/appendix}



\end{document}

%% file: sections/introduction.tex
\section{Introduction}
\label{sec:introduction}

Language models can now solve problems at the level of elite competitive mathematics and programming, often rivaling or surpassing human experts under benchmarked conditions~\cite{huang2025winning, el2025competitive}. This improvement cannot be accounted for by pre-training scale alone and instead reflects deeper gains in multi-step deliberation and tool use. A key driver of this progress is RL fine-tuning, particularly RL with verifiable rewards (RLVR), which shifts models toward longer “thinking” and iterative refinement of intermediate steps. As a result, RL fine-tuning has become a central ingredient in turning strong pretrained models into high-performing reasoning systems.

For example, when DeepSeek-R1 introduced GRPO-style RL training, it reported RL compute of roughly 3\% of pre-training~\cite{khatri2025art}, already meaningful given the gains in reasoning behavior. Since then, as RLVR methods have repeatedly delivered large improvements in multi-step reasoning, there has been growing momentum toward scaling RL post-training, making its efficiency increasingly consequential for end-to-end frontier training.

This shift matters because RL fine-tuning has an entirely different compute profile than pre-training. Unlike pre-training, where compute is dominated by dense forward and backward passes, modern RL fine-tuning spends a large fraction of wall-clock time in rollout generation~\cite{HybridFlow}. 
In RLVR training, the model must generate multiple completions per prompt before rewards and group-relative advantages can be computed. 
This makes group inference, rather than the optimizer step alone, a major contributor to end-to-end training cost. As reasoning models produce longer and more variable chains of thought, this cost becomes even more pronounced. Previous works have reported, group inference to be responsible for 80\% of latency~\cite{qin2025seer}.

  \begin{figure*}[t]
  \centering
  \begin{subfigure}[t]{0.4\textwidth}
    \centering
    \includegraphics[width=\linewidth]{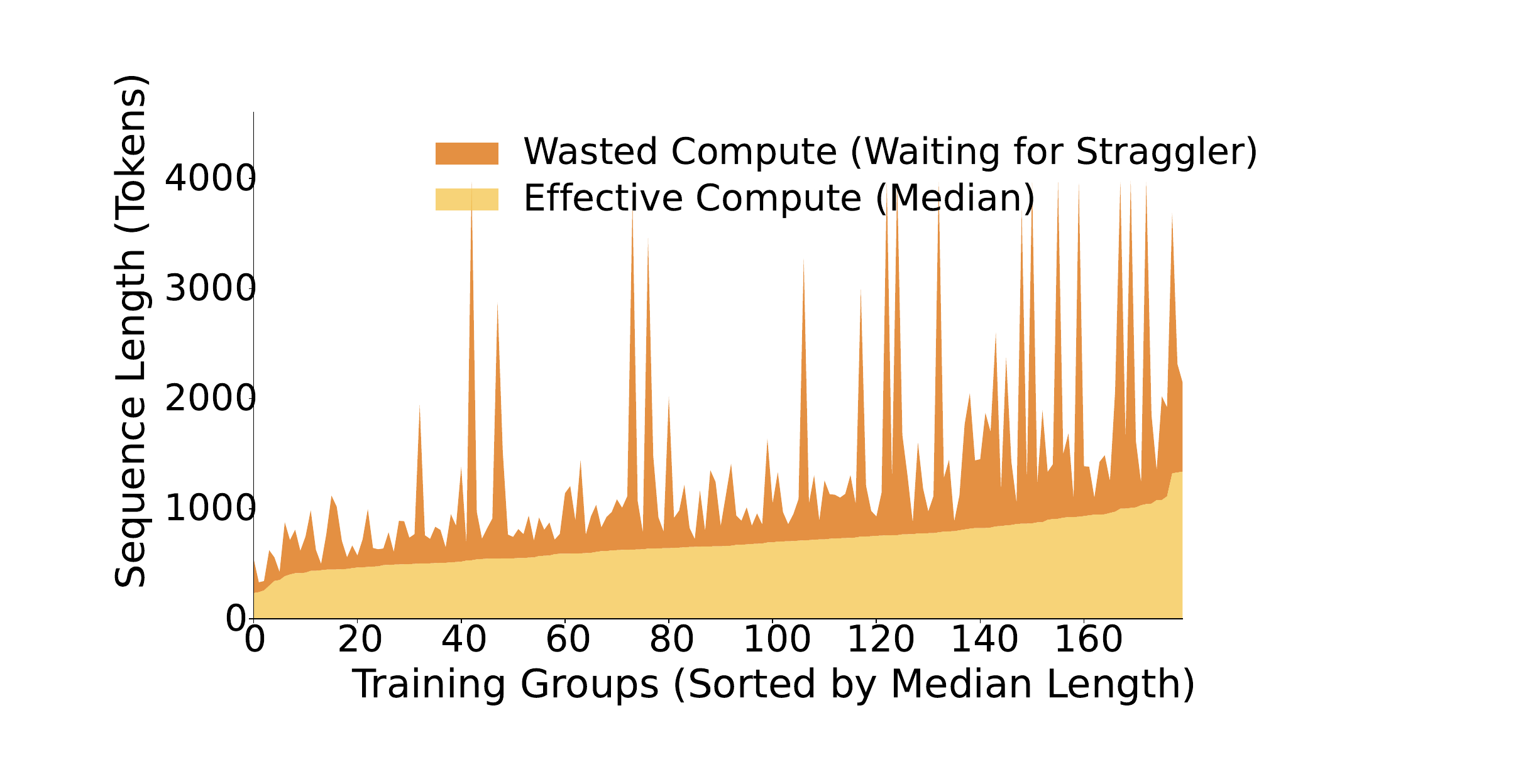}
  \end{subfigure}
  \hspace{-2em}
  \begin{subfigure}[t]{0.32\textwidth}
    \centering
    \includegraphics[width=\linewidth]{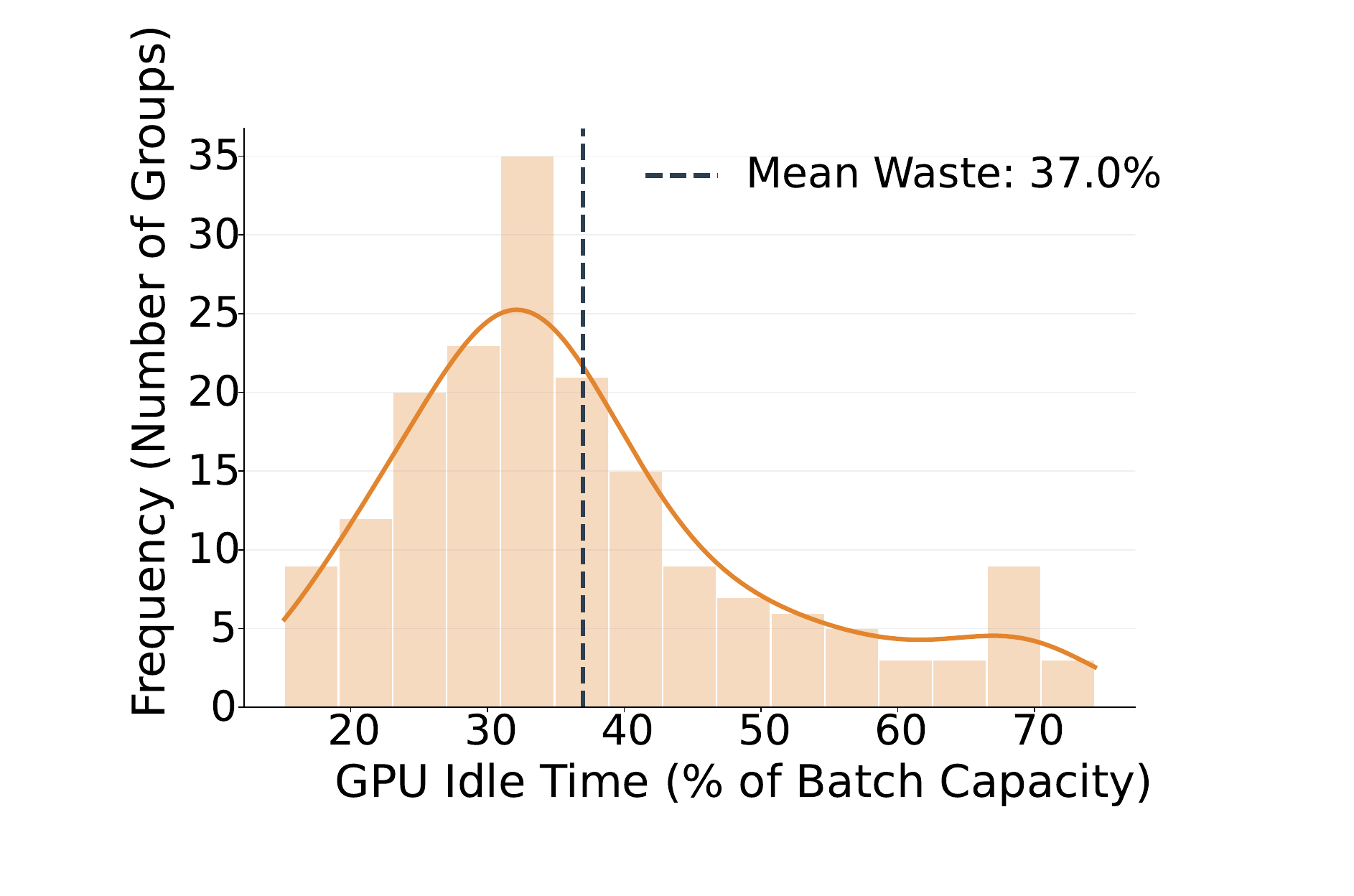}
  \end{subfigure}
  \begin{subfigure}[t]{0.25\textwidth}
    \centering
    \includegraphics[width=\linewidth]{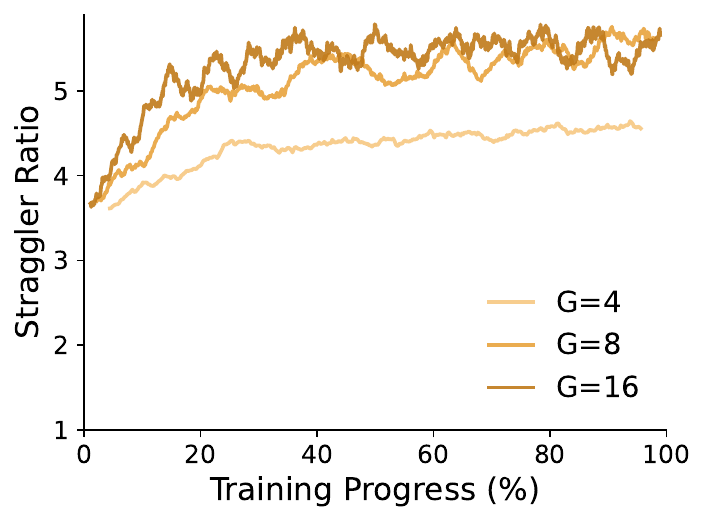}
  \end{subfigure}
  \caption{\textbf{Empirical results of the straggler problem in synchronous RL.} \textbf{Left:} for training groups sorted by median rollout length, the gap between the median and maximum completion length represents GPU time wasted waiting for the slowest rollout. \textbf{Middle:} the distribution of idle-time fraction across groups shows that this wasted capacity is frequent. \textbf{Right:} the straggler ratio (max/mean response length) increases throughout training for all group sizes, showing that the straggler problem worsens as training progresses.}
  \label{fig:motivation}
  \vspace{-1em}
\end{figure*}

Because group inference dominates the cost of RL fine-tuning, the choice of training regime becomes a first-order systems decision. Synchronous RL is attractive because it is strictly on-policy, meaning all rollouts in an update are produced by the same policy snapshot and used immediately, avoiding policy lag and maintaining stable, reproducible training behavior. However, this same synchronization introduces a serious systems bottleneck. In a synchronous group, update time is determined not by the average rollout but by the slowest one. A single unusually long completion can delay reward computation and the backward pass for the entire group, leaving shorter rollouts waiting and wasting hardware capacity (Figure~\ref{fig:motivation}). This \emph{straggler problem} becomes more severe as group size increases, since larger groups are more likely to contain at least one extreme-length rollout.

This effect intensifies exactly in the setting we care about most, scaling RL post-training for reasoning. Longer-horizon deliberation increases both the mean and the variance of rollout lengths, making stragglers more frequent and more severe. As a result, naively increasing group size to improve the statistical quality of advantage estimates can decrease throughput, causing synchronous RL to become progressively less hardware-efficient as reasoning difficulty scales.

This creates a central tension in synchronous RL. Larger group sizes are often desirable because they provide more within-group comparisons and can improve the quality of group-relative updates. But they also increase the likelihood of synchronization stalls and wall-clock inefficiency. A natural alternative is asynchronous training~\cite{afanasyev2026slime}, which improves utilization by relaxing iteration barriers, but this comes at the cost of policy lag and weaker training consistency~\cite{qin2025seer, zhou2026efficient}. Existing systems improvements can reduce overhead substantially e.g., continuous batching, treat requests as independent and do not address the group-level dependency imposed by RL advantage computation. However, these methods do not remove the group-level dependency at the heart of synchronous RL: reward computation and policy updates must still wait for all completions in a group to finish. The question, then, is how to retain the benefits of synchronous, on-policy training while reducing the straggler cost induced by rollout-length variability.


\begin{figure*}
    \centering
    \includegraphics[width=\linewidth]{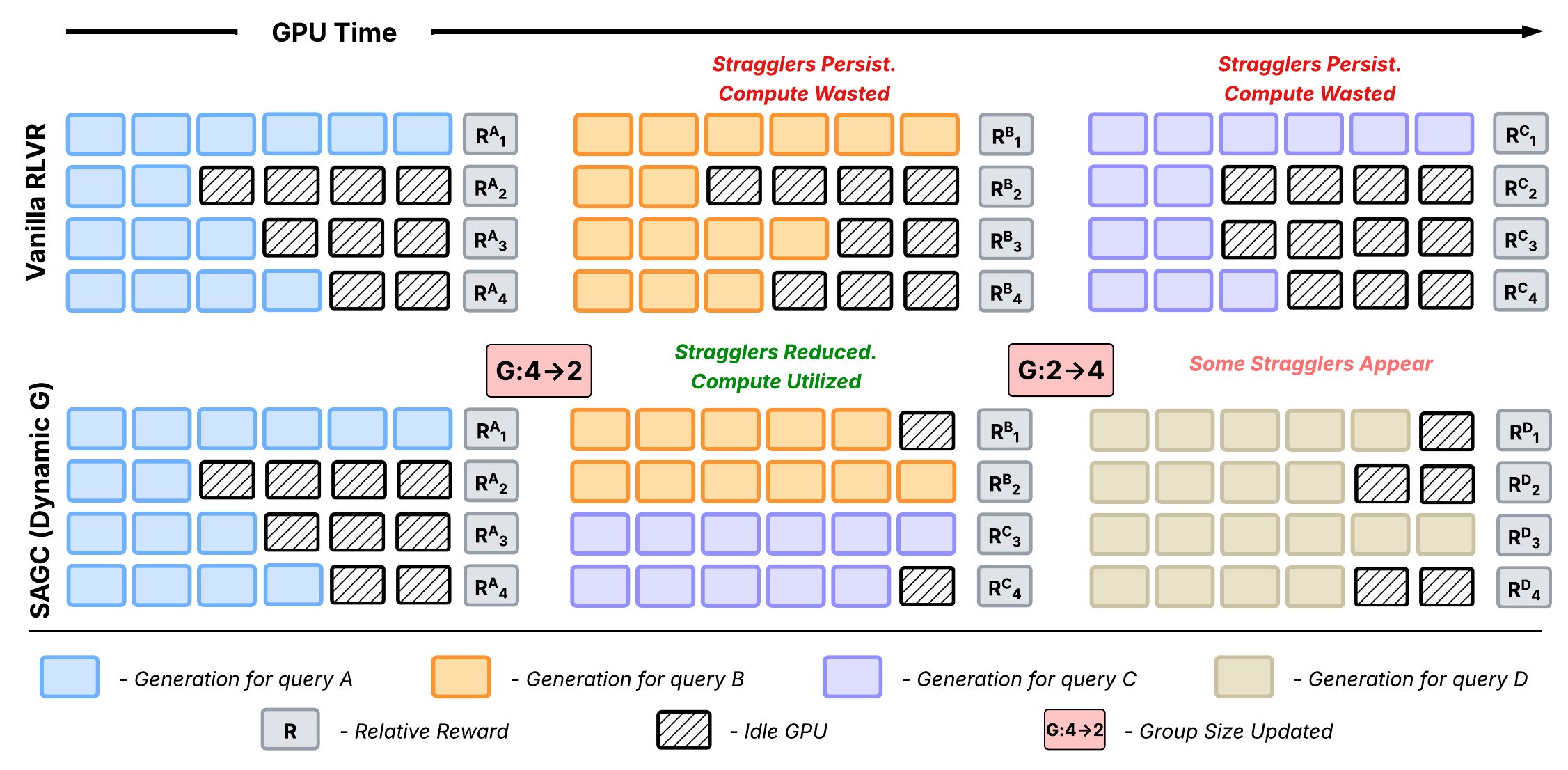}
    \caption{\textbf{Fixed-group synchronous RL wastes hardware efficiency, while \SAGC adapts group size to reduce synchronization stalls.} In vanilla RLVR \textbf{(top)}, a fixed group size $G{=}4$ leads to repeated stragglers, so shorter rollouts wait for the slowest one before rewards and updates can be computed. In \SAGC \textbf{(bottom)}, $G$ is the number of rollouts per query and annotations such as $G{:}4\!\rightarrow\!2$ indicate a controller update of group size for the following queries. By reducing $G$ when rollout variability is high and restoring it when conditions improve, \SAGC shortens synchronization stalls and better utilizes compute.}
    \label{fig:gpu_utilization}
    \vspace{-0.5em}
\end{figure*}

To address this problem, we propose \textbf{S}traggler-\textbf{A}ware \textbf{G}roup \textbf{C}ontrol (\SAGC), a dynamic controller that adapts the training group size online based on observed rollout behavior. Rather than fixing the group size for the entire run, \SAGC monitors group-level length statistics and detects when straggler events occur. It then adjusts future group-size choices to balance the balance the benefit of larger groups against the risk of synchronization stalls.
The key idea is to treat straggler formation as a probabilistic event whose likelihood depends on group size and recent generation statistics. By learning this relationship during training, the controller can proactively avoid configurations that are likely to cause synchronization stalls while still exploiting larger groups when variance is low. This re-frames group sizing from a static hyperparameter into an adaptive control problem. As illustrated in Figure~\ref{fig:gpu_utilization}, in fixed-group synchronous RL a single long rollout can delay reward computation and the backward pass for the entire group, creating large GPU bubbles. Our method reduces this risk by adapting the group size online.


We evaluate \SAGC on two model families under both GRPO and DAPO training, comparing against static group-size baselines as well as a strong engineered synchronous baseline built with OpenRLHF~\cite{hu2024openrlhf}. The results show a consistent pattern. First, \SAGC reduces straggler incidence and improves wall-clock efficiency while achieving competitive or better training reward. Second, these benefits transfer to final model quality, as \SAGC is competitive with or better than the strongest static group-size setting on downstream reasoning benchmarks. Third, \SAGC often produces shorter outputs on downstream tasks without any explicit length penalty, suggesting that straggler-aware control can implicitly discourage unnecessarily long generations. Together, these results show that dynamic group control is a practical way to make synchronous, on-policy RL faster, more robust, and more effective.

Our contributions can be summarized as follows:
\begin{itemize}
    \item We identify intra-group rollout-length variability as a key bottleneck in synchronous, on-policy RL and highlight the resulting trade-off between larger group sizes and straggler-induced wall-clock inefficiency.
    \item We propose \SAGC a straggler-aware dynamic group-size controller that formulates group selection as an online constrained optimization problem, adapting group size to control long-term straggler incidence while preserving the benefit of larger groups.
    \item We demonstrate that these gains persist beyond training-time metrics, yielding competitive or improved downstream reasoning performance and often shorter outputs without any explicit token-length penalty.
\end{itemize}

%% file: sections/related_work.tex
\section{Related Work}
\label{sec:relatedwork}

\paragraph{Reinforcement Learning with Verifiable Rewards.}
A key inflection point in RLVR was Group Relative Policy Optimization (GRPO) ~\cite{shao2024deepseekmath}, introduced in DeepSeek as a PPO~\cite{schulman2017proximal} based objective that removes the critic and uses group-relative baselines, enabling stable RL fine-tuning in domains with sparse, outcome-based feedback. Recently, reasoning models like DeepSeek-R1~\cite{guo2025deepseek} and OpenAI o1~\cite{jaech2024openai} demonstrated that large-scale RLVR can substantially improve multi-step deliberation. Building on GRPO, DAPO~\cite{yu2025dapo} systematized a practical “long-CoT RL” recipe and released an open RL system at scale, reflecting the broader trend that RLVR is now being treated as a scaling knob rather than a small post-training tweak.

\paragraph{RL Finetuning Regimes.}
RL training regimes vary along two independent axes. The first is \textit{on-policy} vs.\ \textit{off-policy}: and the second is \textit{synchronous} vs.\ \textit{asynchronous}. Relaxing either axis yields gains in training efficiency. \textit{Relaxing synchrony} decouples acting and learning: rollout workers generate continuously while the learner updates in parallel, as in asynchronous GRPO~\cite{noukhovitch2025faster}, or updates are performed against slightly stale snapshots, introducing policy lag and requiring objectives that tolerate staleness. \textit{Relaxing on-policy constraints} stays synchronous while reducing long-tail waste through 
rollout repacking~\cite{gao2025rollpacker}, and speculative decoding or stage fusion~\cite{10.5555/3767955.3767981} often sacrificing strict on-policy iteration consistency.

This leaves a gap: \textbf{reducing staleness within a strictly on-policy, synchronous regime}. Our work targets precisely this setting by preserving the algorithmic semantics of synchronous RLVR while eliminating the straggler-induced stalls from intra-group length variability. A complete discussion of related work is provided in Appendix~\ref{sec:relatedworks}.

%% file: sections/motivation.tex
\section{Motivation}
\label{sec:motivation}

\subsection{Problem Statement}
\label{sec:problem_statement}
GRPO trains a policy by sampling multiple responses per prompt and forming relative advantages within them. Concretely, let \(\mathcal{G} = \{4, 8, 16 , \dots\}\) be the set of admissible group sizes, and for each prompt \(x\) we sample a group size \(\G \in \mathcal{G}\) and generate \(\G\) completions \(\{y_i\}_{i=1}^{\G}\), compute rewards \(\{r_i\}_{i=1}^{\G}\), and use group-relative baselines (e.g., mean/standardized rewards) to construct the policy-gradient update. Increasing \(\G\) typically improves the quality and stability of the update by providing more within-group comparisons.

However, batched generation on GPU/TPU is often bottlenecked by the longest completion in the group: shorter sequences incur padding/idle time while waiting for the straggler to finish. Let \(\ell_i\) be the generated token length of completion \(y_i\) (under a fixed maximum \(L_{\max}\)). We define a \emph{straggler ratio} as

{\small
\begin{equation}
R \;=\; \frac{\max_i\, \ell_i}{\operatorname{median}_i\, \ell_i},
    \quad i \in \{1, \dots, \G\}
\label{eq:ratio}
\end{equation}
}
and the corresponding \emph{straggler event}
{\small
\begin{equation}
\hspace{-4.1em} S \;=\; \mathbb{I}[R > \tau], \quad \quad \tau>1
\label{eq:event}
\end{equation}
}


Intuitively, \( S \geq 1 \) implies at least one completion is substantially longer than the typical completion in the group, introducing significant padding inefficiency.

This yields a latency-quality trade-off as larger \(\G\) can improve the 
policy update but increases the likelihood and cost of stragglers. Our proposed approach \emph{adapts \(\G\) online} to reduce GPU idle time caused by straggler-induced staleness by keeping the straggler ratio below a desired target.

\subsection{Constrained Optimization View}
\label{subsec:optimization}

At each training step $t$, choosing a larger \(\G\) improves the statistical 
efficiency of the group-relative update but may increase the frequency of stragglers. We model this as a constrained online decision problem where we select 
\(\G_t \in \mathcal{G}\) to maximize group size while keeping the straggler 
ratio below a desired threshold


{\small
\begin{align}
\max_{\{\G_t\}} \quad & u(\G_t) \\ 
\text{s.t.} \quad & \mathbb{E}[S_t \mid \G_t] \le \delta,
\end{align}
}
where \(\delta\in(0,1)\) is a target straggler rate and \(u(\G)\) is an increasing function of \(\G\). In practice we use \(u(\G)=\log_2(\G)\), which prevents the reward term from dominating the constraint penalty at larger \(\G\) and keeps the utility on a comparable scale to probabilities.

Introducing a Lagrange multiplier \(\lambda \ge 0\), we obtain the online Lagrangian objective:

{\small
\begin{equation}
\mathcal{L}(\G;\lambda) \;=\; u(\G) - \lambda \, p(\G),
\end{equation}
}
where \(p(\G)\approx \mathbb{E}[S_t\mid \G]\) is the unknown, non-stationary straggler probability for group size \(\G\). This reframes dynamic group-size selection as learning \(p(\G)\) online while adjusting \(\lambda\) to enforce the constraint.

\paragraph{Why adapt \(\G\)?}
The key observation is that the marginal benefit of increasing \(\G\) is not constant throughout training, while the marginal throughput cost from stragglers can vary significantly with the model's evolving length distribution.
Early in training, policies often exhibit higher variance in completion length. Later, they may become more concise or more consistently formatted, changing both $\mathbb{E}[S_t \mid \G]$ and the expected padding waste.
As a result, a fixed group size can be suboptimal: choosing a large \(\G\) may waste compute when stragglers are frequent, whereas choosing a small \(\G\) may leave statistical efficiency on the table when stragglers are rare.
This motivates a lightweight \emph{online controller} that adjusts \(\G\) dynamically to satisfy a target straggler rate while preserving the benefit of within-group comparisons.

%% file: sections/method.tex
\section{Methodology: Straggler-Aware Group-Size Control}
\label{sec:method}

We present \textbf{\underline{S}traggler-\underline{A}ware \underline{G}roup-Size \underline{C}ontrol (\SAGC)}, an algorithm-agnostic controller for selecting the group size ($\G$) during training.
\SAGC is compatible with GRPO, DAPO, and related methods that (i) sample multiple completions per prompt, (ii) compute per-completion rewards, and (iii) form group-relative advantages or other within-group statistics.

\subsection{Setup and objective}
\label{subsec:setup_objective}


The goal is to choose \(\G_t\) online to maximize a monotone utility $u(\G)$ over group size, subject to keeping the straggler rate below a target $\delta \in (0,1)$.
We use $u(\G)=\log_2(\G)$, which captures diminishing returns in $G$ and keeps the utility scale comparable across choices.

\subsection{Online Lagrangian and control rule}
\label{subsec:lagrangian_control}

We optimize the constraint using an online primal--dual approach. Introduce a Lagrange multiplier $\lambda_t \ge 0$ and define the per-step Lagrangian score:
\begin{equation}
{\small
\mathcal{L}_t(\G) \;=\; u(\G) - \lambda_t \,\widehat{p}_{t(\G)}
\label{eq:lagrangian_score}
}
\end{equation}
where $\widehat{p}_{t(\G)}$ estimates the non-stationary straggler probability for choosing group size $G$.

\paragraph{Group-size selection.}
At each step, \SAGC chooses the group size via

{\small
\begin{equation}
\G_t \in \arg\max_{\G \in \mathcal{G}} \; \mathcal{L}_t(\G)
\;=\;
\arg\max_{\G \in \mathcal{G}} \left[u(\G) - \lambda_t \widehat{p}_t(\G)\right]
\label{eq:choose_G}
\end{equation}
}
Intuitively, larger $\lambda_t$ penalizes straggler-prone group sizes more strongly.

\paragraph{Dual update.}
After executing step $t$ and observing the straggler event $S_t$, we update $\lambda_t$ by projected dual ascent:

{\small
\begin{equation}
\lambda_{t+1} \;=\; \left[\lambda_t + \eta_\lambda\,(S_t - \delta)\right]_+
\label{eq:lambda_update}
\end{equation}
}
where $\eta_\lambda > 0$ is a step size and $[\cdot]_+$ denotes projection onto $\mathbb{R}_{\ge 0}$.
When stragglers occur more frequently than desired ($S_t > \delta$) on average, $\lambda_t$ increases and the controller becomes more conservative, otherwise $\lambda_t$ decreases, allowing larger \(G\).

\subsection{Estimating straggler probability}
\label{subsec:estimating_p}



For each group size \(\G\), we model the straggler event \(S_t \in \{0,1\}\) as a Bernoulli random variable with unknown probability \(p_{\G}=\Pr(S_t=1\mid \G)\). We place a \(\mathrm{Beta}(\alpha_{\G},\beta_{\G})\) prior over \(p_{\G}\) because Beta is the conjugate prior to the Bernoulli likelihood, yielding simple posterior updates from observed straggler outcomes. Sampling \(\tilde{p}_{\G} \sim \mathrm{Beta}(\alpha_{\G},\beta_{\G})\) therefore gives a posterior sample of the straggler risk for arm \(\G\), which naturally enables Thompson sampling in our controller.

Because the length distribution evolves over training, at each step we update the discounted $\text{Beta}(\alpha_t(\G),\beta_t(\G))$ for the current $\G$ according to Line~11 of \SAGC Algorithm~\ref{alg:sagc}
where $\gamma$ is a forgetting factor that down-weights older observations. At each step, we draw $\tilde{p}_t(\G) \sim \text{Beta}(\alpha_t(\G), \beta_t(\G))$ and select the group size maximizing the sampled Lagrangian score. The probability that any candidate $\G$ is selected is proportional to the posterior probability that it is optimal, naturally balancing exploration of poorly characterized group sizes against exploitation of well-characterized ones. As evidence accumulates, the posterior variance shrinks and the selection concentrates on the group size that best satisfies the constraint in expectation. Finally, the local neighborhood restriction $\mathcal{N}(\G_{t-1})$ in Lines~16--17 of Algorithm~\ref{alg:sagc} ensures the group size selection and the dual variable $\lambda_t$ adapt at comparable rates.

\subsection{Controller implementation and system design}
\label{subsec:controller_impl}

\begin{figure}
    \centering
    \includegraphics[width=0.7\linewidth]{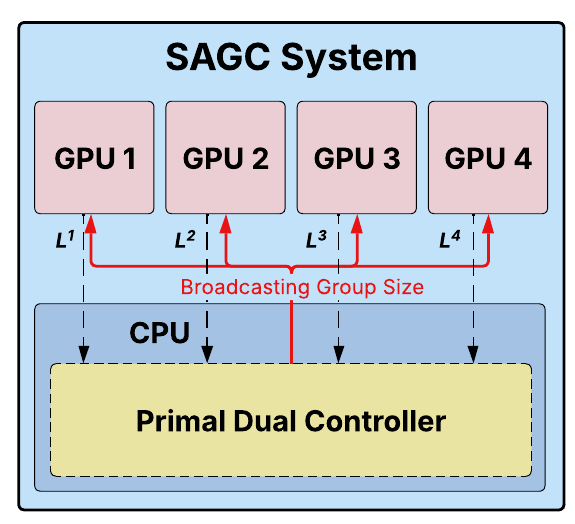}
    \caption{\textbf{System design of Straggler-Aware Group-Size Control (\SAGC).} Each GPU reports lightweight rollout-length statistics to a CPU-side primal-dual controller, which estimates straggler risk, updates the dual variable, and broadcasts the next group size before the following rollout step.}
    \label{fig:system}
    \vspace{-1em}
\end{figure}

\SAGC is implemented as a CPU-side controller that communicates with the training loop with negligible overhead. Figure~\ref{fig:system} shows the control loop used by \SAGC. GPUs emit lightweight length summaries, the CPU-side controller updates the straggler-risk estimate and dual variable, and the selected group size is broadcast before the next rollout step.

During rollout generation at each time step $t$, the GPU computes per-sequence token counts $\ell_i$. For each group we compute summary statistics \(\ell_{\max}, \ell_{\mathrm{median}} \text{ and }\) needed for $R_t$. Only these values are transferred to the CPU.


The CPU controller operates through Algorithm~\ref{alg:sagc} receives $(R_t, S_t, \G_t)$, updates $\lambda_t$ (Line~10), posterior (Line~11) and then selects $G$ for next step $t+1$ (Line~17).
The next group size is then sent back to the training process before the next rollout generation begins, as shown in Figure~\ref{fig:system}.

Crucially, \SAGC does not require gradients, rewards, or model internals. It only needs lengths. 
This makes it a drop-in component for GRPO/DAPO-style pipelines and keeps the control loop inexpensive:
the controller operates on scalars, runs on CPU, and introduces no additional GPU synchronization beyond reporting the already-available length statistics.

\algrenewcommand\algorithmiccomment[1]{\hfill{\color{blue}// #1}}

\begin{algorithm}[t]
\small
\caption{\textbf{SAGC: Straggler-Aware Group-Size Control}}
\label{alg:sagc}
\begin{algorithmic}[1]
\Require Allowed sizes \(\mathcal{G}\) (sorted), target straggler rate \(\delta\), threshold \(\tau\),
dual step size \(\eta_\lambda\), forgetting \(\gamma\in(0,1]\), utility \(u(\G)=\log_2(\G)\),
randomly initialize \(\G_0\in\mathcal{G}\)
\State Initialize \(\lambda \gets 0\)
\ForAll{\(\G\in\mathcal{G}\)} \State \(\alpha_{\G} \gets 1,\ \beta_{\G} \gets 1\) \EndFor
\State \(\G \gets \G_0\)
\While{training}
    \State Run rollout generation with group size \(\G\); collect completed groups
    \State For each completed group: get \((R, S)\) 
    \State \(\bar{S} \gets \text{mean}(S)\) \Comment{Empirical straggler rate over recent groups}
    \State \(\lambda \gets \max\{0,\ \lambda + \eta_\lambda(\bar{S}-\delta)\}\) \Comment{Dual update}

    \State \(\alpha_{\G} \gets \gamma\,\alpha_{\G} + \bar{S}\); \(\beta_{\G} \gets \gamma\,\beta_{\G} + (1-\bar{S})\) 

    \ForAll{\(G \in\mathcal{G}\)}
        \State Sample \(\tilde{p}_G \sim \mathrm{Beta}(\alpha_G,\beta_G)\) 
        \State \(\tilde{U}(G) \gets u(G) - \lambda\,\tilde{p}_G\)
    \EndFor

    \State \(\mathcal{N}(\G) \gets \{\G\}\cup\{\mathrm{prev}(\G),\mathrm{next}(\G)\}\cap\mathcal{G}\) \Comment{Local search for stability}
    \State \(\G \gets \arg\max_{G\in\mathcal{N}(\G)} \tilde{U}(G)\) (tie-break toward smaller \(G\))
\EndWhile
\end{algorithmic}
\end{algorithm}





%% file: sections/experiments.tex
\section{Experiments}
\label{sec:Experiments}

\subsection{Experimental Settings}
\label{sec:experimental_settings}

We evaluate \SAGC on two base language models, Qwen2.5-3B-Instruct and Llama-3.2-3B-Instruct, under GRPO and DAPO training. In all runs, we fix the total training prompt budget $P$ by training on the same prompt set from \texttt{open-r1/DAPO-Math-17k-Processed}~\cite{yu2025dapo}. The controller and all static baselines use the same candidate group sizes, \(\mathcal{G} \in \{4,8,16\}\). For training-time comparisons, we use two systems baselines: Vanilla, a minimally optimized synchronous implementation, and OpenRLHF~\cite{hu2024openrlhf}, a stronger engineered baseline that accelerates training by using vLLM~\cite{vllm} for rollout generation and DeepSpeed for model updates.
Due to the high cost of Vanilla training, we only completed Vanilla runs for the Qwen2.5-3B with GRPO. For the remaining experiments, we use OpenRLHF as the static baseline.

\SAGC is built on top of OpenRLHF, and OpenRLHF vs. \SAGC measures the benefit of dynamic group control beyond systems optimization, while Vanilla vs. OpenRLHF captures the effect of the systems stack. Hyperparameter selection for \SAGC is discussed in Appendix~\ref{sec:implementation_details}. Each run is trained on a single NVIDIA H200 GPU with 141\,GB memory. We report downstream \texttt{pass@4} accuracy on AIME 2023-2025~\cite{aops_aime}, AMC 2023, and GPQA Diamond~\cite{rein2024gpqa}, as well as average response length in generated tokens on the downstream evaluations.

In total, our training and evaluation pipeline consumed approximately 1000 H200 GPU hours, equivalent to more than \$2,400 in compute, spanning 28 training configurations across two model families and two algorithms, along with downstream evaluation on five benchmark suites.

\subsection{How do we ensure a fair comparison between dynamic and static group sizes?}
\label{sec:exp_protocol}
A key challenge in comparing different group sizes is that changing \(\G\) also changes how much generation work is done per optimizer step. Let $\B$ denote the number of prompts processed per step and define the effective generation batch as $\EB = \B \times \G$. If $\B$ is held fixed, then larger \(\G\) automatically receives a larger generation budget per step, which confounds the benefit of dynamic control with the trivial benefit of doing more rollout work. To avoid this, we fix the effective batch at $\EB=64$ for all methods.

Under this protocol, the static baselines with \(\G \in \{4,8,16\} \text{ use } \B \in \{16,8,4\}\) respectively, and \SAGC applies the same rule online with $\B_t = 64/\G_t$. An important consequence is that the number of optimizer steps is not identical across group sizes: for fixed $P$, larger \(\G\) implies smaller $B$ and therefore more training steps. We adopt this normalization because it better reflects a realistic high-utilization setting. All methods operate under the same per-step generation budget, so any wall-clock gain from \SAGC must come from a better execution schedule rather than from processing more generations per step. A complete discussion of fair comparison is provided in Appendix~\ref{app:normalization_choices}

\input{figures/training_tables}

\subsection{How does dynamic group size affect training-time metrics?}
\label{sec:training_metrics}

Table~\ref{tab:training_metrics_combined} first shows the importance of the systems stack itself. Where available, OpenRLHF is dramatically faster and less straggler-prone than the Vanilla baseline. For example, on Qwen2.5-3B with GRPO, moving from Vanilla to OpenRLHF reduces training time from 23.91h to 7.77h at $\G=8$, and from 47.87h to 16.49h at $\G=16$, while also sharply lowering straggler rate. 
The straggler percentage is the fraction of generation groups where the longest response is more than 1.25$\times$ the median response length.
These results establishes OpenRLHF as a strong static baseline.

On top of this optimized baseline, \SAGC consistently improves the training-time trade-off between reward, straggler rate, and wall-clock time. On Qwen2.5-3B with GRPO, \SAGC achieves the best reward ($0.648$) while keeping training time close to the strong OpenRLHF ($\G=8$)  baseline (7.93h vs.\ 7.77h) and substantially reducing straggler rate. In the remaining settings, \SAGC typically attains the best or second-best reward while avoiding the severe straggler cost of always using $\G=16$; for instance, on Llama-3.2-3B with DAPO, \SAGC lowers straggler rate from $99.6\%$ to $43.4\%$ and reduces training time from 29.50h to 16.00h relative to fixed $\G=16$, while remaining close in reward. Overall, these results show that adaptive grouping is complementary to systems optimization. Vanilla vs.\ OpenRLHF implementations measures the value of a better training stack, while OpenRLHF vs.\ \SAGC isolates the value of dynamic control.

\input{figures/downstream_task}

\subsection{How does dynamic group size affect the performance of model on downstream task performance?}
\label{sec:downstream_task}

We evaluate final checkpoints on five downstream reasoning benchmarks on AIME (2023-2025), AMC 2023, and GPQA Diamond.
These evaluations are intended to test whether the systems gains of \SAGC translate into useful reasoning behavior at the final checkpoint, rather than merely improving intermediate training-time metrics such as straggler rate or wall-clock throughput.

Table~\ref{tab:downstream_grpo_dapo_combined} shows that \SAGC is consistently competitive with the strongest static baseline and often matches or outperforms it. The clearest gains appear under DAPO, where \SAGC achieves the best AIME and AMC results for both model families and is also best or tied-best on GPQA. Under GRPO, the results are more mixed, but \SAGC remains close to the best static group while avoiding facing stragglers during the training.

These downstream results should be read together with the training metrics in Table~\ref{tab:training_metrics_combined}. The strongest static downstream numbers often come from larger group sizes, but those settings are also the most straggler-prone and the most expensive in wall-clock time, almost twice as much as \SAGC. \SAGC recovers the benefit of larger groups without paying that cost uniformly across training, yielding a better quality-efficiency trade-off overall.

\input{figures/downstream_length}

\subsection{How does dynamic group size affect the length of outputs?}
\label{sec:downstream_length}

Table~\ref{tab:downstream_length_grpo_dapo_combined} reports average response length on downstream benchmarks. We find that \SAGC often produces shorter responses while maintaining, and in several cases improving, downstream accuracy. This trend is especially visible on GPQA and AMC, where \SAGC is frequently among the shortest methods while remaining competitive with or better than the strongest static baseline.

Notably, \SAGC does not use any explicit token-level penalty. Instead, the shorter outputs suggest an implicit length penalty induced by the controller: since straggler events are driven by unusually long and high-variance generations, operating points that repeatedly produce such rollouts become less attractive during training. In this sense, \SAGC biases training away from excessively long generations without directly modifying the reward, helping reduce unnecessary verbosity while preserving useful reasoning behavior.

\subsection{Analysis of Posterior Risk Estimates}
\label{sec:posterior}

\begin{figure}[H]
    \centering
    \includegraphics[width=0.4\textwidth]{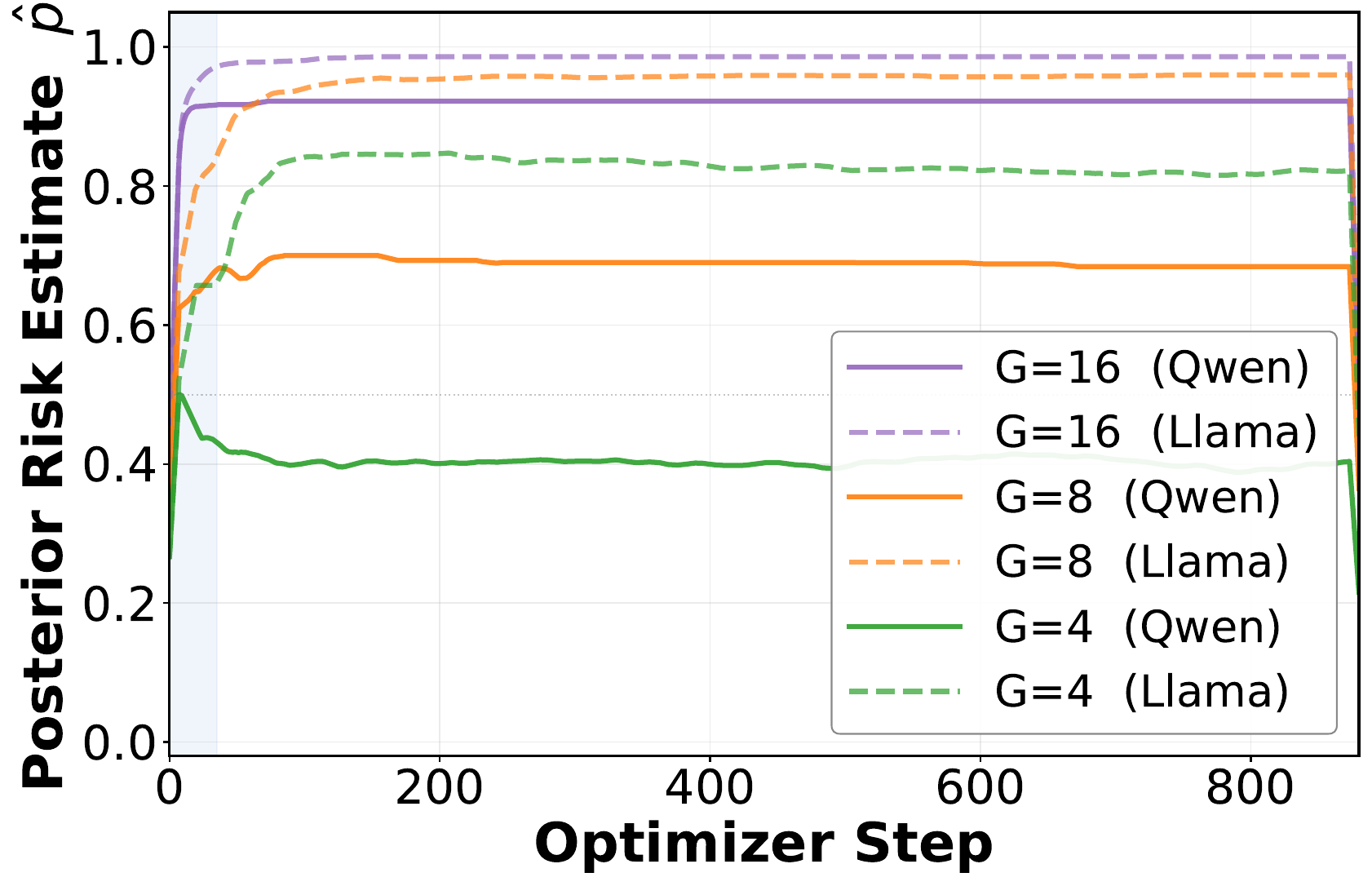}
    \caption{\textbf{Posterior risk estimates for each candidate group size (G $\in$ {4, 8, 16}) over 882 optimizer steps.} Higher values indicate greater straggler probability for that group size.}
    \label{fig:posterior_risk}
\end{figure}

\vspace{-1em}

Figure~\ref{fig:posterior_risk} shows the evolution of the posterior risk estimates ($\hat{p}_G = \frac{\alpha_G}{\alpha_G + \beta_G}$) for each candidate group size (G $\in$ {4, 8, 16}) during training. Starting from a uniform $\mathrm{Beta}(1,1)$ prior, the estimates separate within approximately 30 optimizer steps (shaded region), after which the ordering $(\hat{p}_{16} > \hat{p}_8 > \hat{p}_4)$ is consistently maintained. This confirms that the Bayesian mechanism rapidly identifies the straggler risk profile: larger groups carry higher straggler probability. The posterior estimates are systematically higher for Llama than for Qwen across all arms, reflecting Llama inherently larger response-length variability on mathematical reasoning tasks. This model-level separation demonstrates that the controller captures differences in generation characteristics and adapts its group-size policy accordingly, without manual tuning.

%% file: figures/training_tables.tex
\definecolor{bestcolor}{RGB}{220,252,231}
\definecolor{headercolor}{RGB}{59,130,246}

\begin{table}[H]
\centering
\scriptsize
\setlength{\tabcolsep}{3pt}
\renewcommand{\arraystretch}{1.08}
\caption{Training metrics in one-column format. Blocks denote model--algorithm pairs. Metric abbreviations are: \textbf{T} = training time in hours, \textbf{Str} = straggler percentage, \textbf{RW} = final reward, and \textbf{Len} = average response length in generated tokens. Best within each block is \textbf{bold}; second-best is \underline{underlined}. $\uparrow$ indicates higher is better and $\downarrow$ indicates lower is better.}
\label{tab:training_metrics_combined}

\resizebox{\columnwidth}{!}{%
\begin{tabular}{llcccc}
\toprule
\rowcolor{headercolor!15}
\textbf{Block} & \textbf{Configuration} & \textbf{T$\downarrow$} & \textbf{Str$\downarrow$} & \textbf{RW$\uparrow$} & \textbf{Len$\downarrow$} \\
\midrule

\multirow{7}{*}{\shortstack{\textbf{Qwen}\\\textbf{GRPO}}}
& Vanilla (G=4)   & 15.79 & 26.9 & 0.063 & 758 \\
& Vanilla (G=8)   & 23.91 & 59.8 & 0.098 & 746 \\
& Vanilla (G=16)  & 47.87 & 87.8 & 0.117 & 751 \\
& OpenRLHF (G=4)  & \textbf{4.05} & \textbf{3.9} & 0.582 & \underline{733} \\
& OpenRLHF (G=8)  & \underline{7.77} & 14.1 & \underline{0.594} & \underline{733} \\
& OpenRLHF (G=16) & 16.49 & 31.4 & 0.461 & \textbf{727} \\
\rowcolor{bestcolor}
& \textbf{SAGC (Dynamic)} & 7.93 & \underline{8.5} & \textbf{0.648} & 735 \\
\midrule

\multirow{4}{*}{\shortstack{\textbf{Qwen}\\\textbf{DAPO}}}
& OpenRLHF (G=4)  & \textbf{3.52} & \textbf{39.9} & 0.083 & 736 \\
& OpenRLHF (G=8)  & \underline{6.80} & 66.7 & 0.112 & \textbf{730} \\
& OpenRLHF (G=16) & 13.28 & 88.1 & \textbf{0.132} & 735 \\
\rowcolor{bestcolor}
& \textbf{SAGC (Dynamic)} & 10.81 & \underline{41.6} & \underline{0.121} & \underline{731} \\
\midrule

\multirow{4}{*}{\shortstack{\textbf{Llama}\\\textbf{GRPO}}}
& OpenRLHF (G=4)  & \textbf{9.27} & \underline{63.6} & 0.438 & 959 \\
& OpenRLHF (G=8)  & \underline{17.70} & 88.7 & 0.527 & \underline{803} \\
& OpenRLHF (G=16) & 34.99 & 95.4 & \textbf{0.742} & \textbf{740} \\
\rowcolor{bestcolor}
& \textbf{SAGC (Dynamic)} & 18.00 & \textbf{58.7} & \underline{0.566} & 849 \\
\midrule

\multirow{4}{*}{\shortstack{\textbf{Llama}\\\textbf{DAPO}}}
& OpenRLHF (G=4)  & \textbf{8.27} & 84.0 & 0.062 & 969 \\
& OpenRLHF (G=8)  & \underline{15.73} & 97.1 & 0.075 & 858 \\
& OpenRLHF (G=16) & 29.50 & 99.6 & \textbf{0.088} & \textbf{734} \\
\rowcolor{bestcolor}
& \textbf{SAGC (Dynamic)} & 16.00 & \textbf{43.4} & \underline{0.077} & \underline{840} \\
\bottomrule
\end{tabular}%
}
\end{table}

%% file: figures/downstream_task.tex
\definecolor{headercolor}{RGB}{59,130,246}
\definecolor{bestcolor}{RGB}{220,252,231}

\begin{table}[H]
\centering
\scriptsize
\setlength{\tabcolsep}{3.5pt}
\renewcommand{\arraystretch}{1.08}

\caption{Experimental results of trained models on downstream task accuracy (pass@4, \%). \textbf{AIME(avg)} is the macro-average over AIME 2023/2024/2025, \textbf{AMC} denotes AMC 2023, and \textbf{GPQA} denotes GPQA Diamond. Best within each block is \textbf{bold}; second-best is \underline{underlined}.}
\label{tab:downstream_grpo_dapo_combined}

\resizebox{\columnwidth}{!}{%
\begin{tabular}{llccc}
\toprule
\rowcolor{headercolor!15}
\textbf{Block} & \textbf{Configuration} & \textbf{AIME(avg)} & \textbf{AMC 23} & \textbf{GPQA} \\
\midrule

\multirow{4}{*}{\shortstack{\textbf{Qwen2.5-3b}\\\textbf{GRPO}}}
& OpenRLHF (G=4)   & \underline{6.67} & 55.00 & \textbf{61.11} \\
& OpenRLHF (G=8)   & \underline{6.67} & \underline{60.00} & 57.07 \\
& OpenRLHF (G=16)  & \textbf{7.78} & \textbf{67.50} & \underline{60.61} \\
\rowcolor{bestcolor}
& \textbf{SAGC (Ours)} & \textbf{7.78} & \underline{60.00} & \textbf{61.11} \\
\midrule

\multirow{4}{*}{\shortstack{\textbf{Qwen2.5-3b}\\\textbf{DAPO}}}
& OpenRLHF (G=4)   & 5.56 & \underline{67.50} & \underline{55.05} \\
& OpenRLHF (G=8)   & 5.56 & 65.00 & \textbf{56.06} \\
& OpenRLHF (G=16)  & \underline{8.89} & 62.50 & \underline{55.05} \\
\rowcolor{bestcolor}
& \textbf{SAGC (Ours)} & \textbf{10.00} & \textbf{75.00} & \textbf{56.06} \\
\midrule

\multirow{4}{*}{\shortstack{\textbf{Llama3.2-3b}\\\textbf{GRPO}}}
& OpenRLHF (G=4)   & 6.67 & 40.00 & 57.07 \\
& OpenRLHF (G=8)   & \underline{7.78} & 35.00 & 57.58 \\
& OpenRLHF (G=16)  & \textbf{8.89} & \textbf{45.00} & \textbf{63.64} \\
\rowcolor{bestcolor}
& \textbf{SAGC (Ours)} & \underline{7.78} & \underline{42.50} & \underline{62.12} \\
\midrule

\multirow{4}{*}{\shortstack{\textbf{Llama3.2-3b}\\\textbf{DAPO}}}
& OpenRLHF (G=4)   & 6.67 & 32.50 & \underline{62.63} \\
& OpenRLHF (G=8)   & 6.67 & 37.50 & \underline{62.63} \\
& OpenRLHF (G=16)  & \underline{10.00} & \underline{42.50} & 61.11 \\
\rowcolor{bestcolor}
& \textbf{SAGC (Ours)} & \textbf{11.11} & \textbf{47.50} & \textbf{67.17} \\
\bottomrule
\end{tabular}%
}
\end{table}

%% file: figures/downstream_length.tex
\definecolor{headercolor}{RGB}{59,130,246}
\definecolor{bestcolor}{RGB}{220,252,231}

\begin{table}[H]
\centering
\scriptsize
\setlength{\tabcolsep}{3.5pt}
\renewcommand{\arraystretch}{1.08}

\caption{Experimental results of trained models on downstream response length (tokens). \textbf{AIME(avg)} is the macro-average over AIME 2023/2024/2025, \textbf{AMC} denotes AMC 2023, and \textbf{GPQA} denotes GPQA Diamond. Lower is better. Best within each block is \textbf{bold}; second-best is \underline{underlined}.}
\label{tab:downstream_length_grpo_dapo_combined}

\resizebox{\columnwidth}{!}{%
\begin{tabular}{llccc}
\toprule
\rowcolor{headercolor!15}
\textbf{Block} & \textbf{Configuration} & \textbf{AIME(avg)} & \textbf{AMC 23} & \textbf{GPQA} \\
\midrule

\multirow{4}{*}{\shortstack{\textbf{Qwen2.5-3b}\\\textbf{GRPO}}}
& OpenRLHF (G=4)   & \textbf{949.8} & 882.8 & 511.1 \\
& OpenRLHF (G=8)   & 969.1 & 912.8 & 499.1 \\
& OpenRLHF (G=16)  & 992.9 & \textbf{839.3} & \underline{481.4} \\
\rowcolor{bestcolor}
& \textbf{SAGC (Ours)} & \underline{960.3} & \underline{880.0} & \textbf{470.0} \\
\midrule

\multirow{4}{*}{\shortstack{\textbf{Qwen2.5-3b}\\\textbf{DAPO}}}
& OpenRLHF (G=4)   & 1025.5 & 833.5 & 497.7 \\
& OpenRLHF (G=8)   & 966.3 & 839.2 & 495.2 \\
& OpenRLHF (G=16)  & \textbf{931.4} & \underline{817.3} & \underline{491.9} \\
\rowcolor{bestcolor}
& \textbf{SAGC (Ours)} & \underline{936.3} & \textbf{803.0} & \textbf{472.2} \\
\midrule

\multirow{4}{*}{\shortstack{\textbf{Llama3.2-3b}\\\textbf{GRPO}}}
& OpenRLHF (G=4)   & 1079.2 & \textbf{850.2} & 451.1 \\
& OpenRLHF (G=8)   & 1068.8 & 915.9 & \underline{422.4} \\
& OpenRLHF (G=16)  & \underline{1052.3} & 992.7 & 435.3 \\
\rowcolor{bestcolor}
& \textbf{SAGC (Ours)} & \textbf{1029.6} & \underline{887.9} & \textbf{414.8} \\
\midrule

\multirow{4}{*}{\shortstack{\textbf{Llama3.2-3b}\\\textbf{DAPO}}}
& OpenRLHF (G=4)   & 1155.9 & 933.0 & 464.2 \\
& OpenRLHF (G=8)   & 1123.5 & 927.7 & \textbf{432.5} \\
& OpenRLHF (G=16)  & \underline{1057.2} & \underline{882.2} & \underline{433.8} \\
\rowcolor{bestcolor}
& \textbf{SAGC (Ours)} & \textbf{1049.8} & \textbf{838.7} & 463.9 \\
\bottomrule
\end{tabular}%
}
\end{table}

%% file: sections/conclusion.tex
\section{Conclusion}
\label{sec:conclusion}
We presented Straggler-Aware Group Control (\SAGC), a lightweight controller that adapts group size online to reduce synchronization stalls in synchronous on-policy RL. Across GRPO and DAPO, and across two model families, \SAGC consistently improved the trade-off between wall-clock efficiency, straggler rate, and final training reward. These gains also transferred to downstream reasoning benchmarks. Overall, our results show that dynamic group control is a practical and effective systems lever for making synchronous RL training faster and more robust.
Additional details are provided in the Appendix.

%% file: sections/limitations.tex
\section*{Limitations}
A limitation of the current controller is that it is intentionally independent of prompt type or prompt class. \SAGC only uses observed rollout-length statistics, which keeps it lightweight, annotation-free, and easy to integrate into existing synchronous RL pipelines. However, this also means it does not exploit prompt-level signals that might help anticipate straggler-prone queries earlier. In principle, one could incorporate prompt-aware difficulty or length predictors, but doing so reliably is challenging in the on-policy setting because the policy itself changes throughout training, and practical proxy approaches may require annotated data or additional modeling assumptions. Exploring hybrid controllers that preserve the simplicity of \SAGC while incorporating such prompt-aware priors is a promising direction for future work.

\section*{Acknowledgments}
The work of Azal Ahmad Khan was supported by the Amazon Machine Learning Systems Fellowship. The work of Ali Anwar was supported by the Samsung Global Research Outreach Award and the National Science Foundation Privacy-Preserving Data Sharing in Practice (PDaSP) program under grant number 2452817.

%% file: sections/appendix.tex
\clearpage
\onecolumn

\appendix

\input{sections/related_works}

\section{Alternative normalization choices for group-size experiments}
\label{app:normalization_choices}

When varying the group size $\G$, there is no unique notion of a ``fair'' comparison unless one specifies what quantity is held fixed. Let $B$ denote the number of prompts processed per optimizer step, and let
\[
EB = B \times \G
\]
denote the effective generation batch. For a fixed training prompt budget $P$, the number of optimizer steps is
\[
S = \frac{P}{B},
\]
and the total number of sampled completions is
\[
T = P \times \G = S \times EB.
\]

Two natural protocols are possible. One may keep $B$ fixed and vary $\G$, in which case the number of optimizer steps stays fixed but the effective generation batch grows with $\G$. Alternatively, one may keep $EB$ fixed and vary $B$ inversely with $\G$, in which case all methods operate under the same per-step generation budget, but the number of optimizer steps depends on the chosen group size. We adopt the second protocol in the main paper.

Table~\ref{tab:appendix_fixed_b} illustrates the alternative fixed-$B$ regime using $B=8$ for all static baselines. Table~\ref{tab:appendix_fixed_eb} shows the fixed-$EB$ regime used in our experiments, where $EB=64$ is held constant across both static baselines and \SAGC.

\begin{table}[t]
\centering
\small
\setlength{\tabcolsep}{6pt}
\renewcommand{\arraystretch}{1.15}
\begin{tabular}{@{}ccccc@{}}
\toprule
\textbf{$\G$} & \textbf{$B$} & \textbf{$EB = B \times G$} & \textbf{$S = P/B$} & \textbf{$T = P \times \G$} \\
\midrule
4  & 8 & 32  & $P/8$ & $4P$  \\
8  & 8 & 64  & $P/8$ & $8P$  \\
16 & 8 & 128 & $P/8$ & $16P$ \\
\bottomrule
\end{tabular}
\caption{Illustration of the \textbf{fixed-$B$} normalization. Here the number of prompts per optimizer step is held constant, so the number of optimizer steps remains unchanged across group sizes. However, larger group sizes automatically receive a larger effective generation batch per step.}
\label{tab:appendix_fixed_b}
\end{table}

\begin{table}[t]
\centering
\small
\setlength{\tabcolsep}{6pt}
\renewcommand{\arraystretch}{1.15}
\begin{tabular}{@{}lcccc@{}}
\toprule
\textbf{Method / Group} & \textbf{$B$} & \textbf{$EB$} & \textbf{Optimizer Steps} & \textbf{Total Generations} \\
\midrule
Static $G=4$   & 16          & 64 & $P/16$             & $4P$              \\
Static $G=8$   & 8           & 64 & $P/8$              & $8P$              \\
Static $G=16$  & 4           & 64 & $P/4$              & $16P$             \\
\textbf{\SAGC (dynamic)} & $B_t = 64/G_t$ & 64 & schedule-dependent & schedule-dependent \\
\bottomrule
\end{tabular}
\caption{The \textbf{fixed-effective-batch} protocol used in the main paper. All methods operate under the same per-step generation budget ($EB=64$), while the number of optimizer steps varies with the chosen group size. For \SAGC, both the prompt batch size and the total number of steps are determined by the learned schedule $\G_t \in \{4,8,16\}$.}
\label{tab:appendix_fixed_eb}
\end{table}

\section{Implementation Details}
\label{sec:implementation_details}

\subsection{Signals used by the controller}
During GRPO training we already generate \(\G\) completions per prompt. We additionally log the per-completion token lengths \(\{\ell_i\}\) and package them into per-prompt groups. Each completed group contributes one straggler observation via the max/median ratio, which is robust to outliers and invariant to absolute scale.

\subsection{Integration into the GRPO training loop}
The controller is implemented as a training callback that executes on each optimizer step:
(i) it drains a FIFO queue of recently completed groups (token-length vectors),
(ii) computes the batch straggler statistics, and
(iii) updates the trainer's \texttt{num\_generations} for subsequent rollouts.
This design keeps the controller asynchronous with respect to generation while ensuring that group-size changes occur at stable boundaries (optimizer steps).

\subsection{Practical considerations}
We restrict actions to neighbor moves in \(\mathcal{G}\) (e.g., \(\{4,8,16\}\)) to prevent oscillations and reduce sensitivity to noisy measurements. We use \(\log_2(\G)\) as the benefit term so that the constraint penalty \(\lambda \tilde{p}_{\G}\) remains influential at larger \(\G\). In systems where generation may return fewer than \(\G\) completions per microbatch (e.g., certain servered generation modes), groups can be assembled across microbatches until the expected group size is reached, then emitted as a single observation for the controller.

\section{Controller Hyperparameters}
The main controller knobs in \SAGC are the straggler threshold \(\tau\) and the target straggler rate \(\delta\). We intentionally chose a conservative setting, since our primary goal was to suppress severe straggler-induced stalls in synchronous RL. Under such a setting, if even smaller groups already exceed the target straggler budget, the controller will naturally avoid larger groups thereafter. We view this as an intended consequence of the constrained design: \SAGC is meant to select the largest group size that remains compatible with stable execution, not to pursue larger groups when the observed rollout distribution is already highly straggler-prone. Relaxing \(\tau\) or increasing \(\delta\) would make the controller more permissive, but our experiments prioritized aggressive straggler reduction because that was the first-order bottleneck in our setting.

%% file: sections/related_works.tex
\section{Related Works}
\label{sec:relatedworks}


\paragraph{Reinforcement Learning with Verifiable Rewards (RLVR).}
RLVR has emerged as a central post-training paradigm for reasoning LLMs, replacing external reward model with programmatic, checkable signals 
A key inflection point was Group Relative Policy Optimization (GRPO) ~\cite{shao2024deepseekmath}, introduced in DeepSeek as a PPO~\cite{schulman2017proximal} based objective that removes the critic and uses group-relative baselines, enabling stable RL fine-tuning in domains with sparse, outcome-based feedback. Recently, reasoning models like DeepSeek-R1~\cite{guo2025deepseek} and OpenAI o1~\cite{jaech2024openai} demonstrated that large-scale RLVR can substantially improve multi-step deliberation. Building on GRPO, DAPO~\cite{yu2025dapo} systematized a practical “long-CoT RL” recipe and released an open RL system at scale, reflecting the broader trend that RLVR is now being treated as a scaling knob rather than a small post-training tweak.

\paragraph{RL Finetuning Regimes.}

RL training regimes vary along two independent axes. The first is \textit{on-policy} vs.\ \textit{off-policy}: whether gradient updates consume rollouts from the current policy snapshot exclusively~\cite{schulman2017proximal, NEURIPS2022_b1efde53}, or tolerate data collected under older policies~\cite{noukhovitch2025faster}. The second is \textit{synchronous} vs.\ \textit{asynchronous}: whether rollout collection and parameter updates alternate in strict lock-step barriers~\cite{hu2024openrlhf}, or overlap continuously via decoupled actor-learner pipelines~\cite{pmlrespeholt18a, noukhovitch2025faster}.

On-policy synchronous training is the strictest quadrant as each iteration collects a full rollout batch, waits for completion and then updates ensuring all rollouts originate from a single policy snapshot but coupling iteration progress to the slowest rollout and causing straggler-induced stalls that worsen as group sizes grow.

A natural reaction is to improve utilization by relaxing either axis. \textit{Relaxing synchrony} decouples acting and learning: rollout workers generate continuously while the learner updates in parallel, as in asynchronous GRPO~\cite{noukhovitch2025faster}, or updates are performed against slightly stale snapshots, introducing policy lag and requiring objectives that tolerate staleness. \textit{Relaxing on-policy constraints} stays synchronous while reducing long-tail waste through length-aware request ordering~\cite{}, rollout repacking~\cite{gao2025rollpacker}, and speculative decoding or stage fusion~\cite{10.5555/3767955.3767981} often sacrificing strict iteration consistency.

\paragraph{Length-Aware Penalties.}
Improvements in RLVR-style reasoning training often come with longer and higher-variance CoT rollouts~\cite{xu2025thinking}, which increase both training-time sampling cost and inference-time latency~\cite{su2025between}. As a result, a growing body of work explicitly incorporates length-aware penalties~\cite{arora2025training}, or formulating constrained objectives that maximize task reward subject to token-budget constraints~\cite{aggarwal2025l}. 

Unlike methods that explicitly penalize long rewards, we implicitly reduce excessively long outputs. Our controller restricts “straggler” rollouts, which are unusually lengthy generations that hinder progress.

\paragraph{Efficient Reasoning.}
Recent work on efficient reasoning can be viewed along two complementary axes: improving \emph{training efficiency} and improving \emph{inference efficiency}. On the inference side, prior work has aimed to reduce the cost of long-form reasoning by controlling or shortening chain-of-thought length, pruning redundant reasoning, or accelerating deliberation with draft-and-verify style mechanisms and speculative reasoning \citep{sui2025stop, sun2024fast, yang2024buffer, cheshmi2025accelerating, ahmed2025retrieval}. On the training side, recent systems and RL works have sought to make reasoning post-training more efficient by improving rollout utilization, stage scheduling, or length-aware optimization during RL itself \citep{nimmaturi2025predictive, wang2025grpo, liu2025prefix, li2025branchgrpo, pang2025theory}. Our work primarily belongs to the training-efficiency axis: rather than changing the test-time decoding procedure or explicitly penalizing long generations, we improve the efficiency of synchronous on-policy RL by adapting group size online to reduce straggler-induced stalls. At the same time, our results suggest that better training-time control can also yield an implicit inference-time benefit, since \SAGC often produces shorter downstream responses without any explicit length penalty.